\documentclass[letterpaper, 10 pt, conference]{ieeeconf}  

\IEEEoverridecommandlockouts                              

\usepackage{graphicx}
\usepackage{listings}
\usepackage{url}

\title{\LARGE \bf
Modeling Non-Functional Application Domain Constraints for Component-Based Robotics Software Systems
}

\newcommand{\smart}{\textsc{Smart\-Soft}}

\author{Alex Lotz$^{1}$, Arne Hamann$^{2}$, Ingo L\"{u}tkebohle$^{2}$,\\ Dennis Stampfer$^{1}$, Matthias Lutz$^{1}$ and Christian
Schlegel$^{1}$
\thanks{$^{1}$A. Lotz, D. Stampfer, M. Lutz and C. Schlegel are with the Faculty of Computer Science,
        University of Applied Sciences, 89075 Ulm, Germany
        {\tt\small \{lotz|stampfer|lutz|schlegel\}@hs-ulm.de}
        }%
\thanks{$^{2}$Arne Hamann and Ingo L\"{u}tkebohle are with Robert Bosch Corporate Research,
        Robert-Bosch-Campus 1, 71272 Renningen, Germany\newline
        {\tt\small \{Arne.Hamann|Ingo.Luetkebohle\}@de.bosch.com}
        }%
}

\begin{document}

\maketitle
\thispagestyle{empty}
\pagestyle{empty}

\begin{abstract}

Service robots are complex, heterogeneous, software intensive systems built from components.
Recent robotics research trends mainly address isolated capabilities on functional level.
Non-functional properties, such as responsiveness or deterministic behavior, are addressed 
only in isolation (if at all).
We argue that handling such non-functional properties on system level is a crucial next step. 
We claim that precise control over application-specific, dynamic execution and interaction behavior of
functional components -- i.e. clear computation and communication
semantics on model level without hidden code-defined parts -- is a key ingredient thereto.

In this paper, we propose modeling concepts for these semantics, and present a meta-model which 
(i) enables component developers to implement component functionalities without presuming application-specific,
system-level attributes, and (ii) enables system integrators to reason about causal dependencies
between components as well as system-level data-flow characteristics. This allows to control data-propagation 
semantics and system properties such as end-to-end latencies during system integration without breaking
component encapsulation.

\end{abstract}

\section{Introduction and Motivation}
\label{sec:introduction}

Service robots are complex, heterogeneous, software intensive systems.
Recent robotics research trends mainly address isolated capabilities on functional level.
Examples include robust perception, mobile manipulation and intuitive human-robot interaction.
This already allows to showcase impressive lab prototypes. 
However, the inter-disciplinary software engineering challenges, i.e. building modular and flexible software architectures covering
several product generations that are easy to maintain and that adhere to functional and in particular non-functional requirements, are underestimated and underrepresented.
The general need for model-based system engineering techniques within the robotics domain is also recognized by the \emph{European SPARC Robotics}~\cite{SPARC}
initiative in its \emph{Multi-Annual Roadmap (MAR)}~\cite{MAR} as ``\emph{the ``make or break'' factor in the development of complex robot
systems}'' -- \cite{MAR}.

Service robotics as a science of integration relies on the combination of individual expertise from various stakeholders involved in the overall development of
a robotics software system. In this paper we particularly focus on two distinct expert roles: the \emph{robotics experts}, i.e. experts of a particular
technology domain such as computer vision, mobile manipulation, etc., and \emph{application experts} for various application domains such as e.g. logistics,
agriculture, etc.

Obviously, both experts need to focus on different concerns to build a robotic system. For instance, \emph{robotics experts} should be able to focus on the
functional part of a component without anticipating application specific details, whereas the \emph{application domain experts} should be able to select the
right components and to adjust them on model level according to the application related requirements without the need to investigate nor modify the internal
implementation of the individual software parts.

We therefore argue that successful and efficient system engineering strongly relies on a clear separation of concerns
allowing for efficient collaboration between all involved stakeholders \cite{Lotz:2014:IJISMD}. Only then the different experts are enabled
to fully concentrate on their dedicated expertise, which globally leads to shared and lowered risks, increased robustness and product quality as well as
reduced costs, development time, and time to the market.

While several model-based robotics approaches such as RTC~\cite{Ando:2005:RTC}, RobotML~\cite{Dhouib:2012:RobotML} and BCM~\cite{Bruyninckx:2013:BCM} already
facilitate the description of functional components by \emph{robotics experts}, the system integration part which is central for \emph{application domain
experts} is currently not systematically addressed in robotics system development (besides of a few promising initiatives such as Rock~\cite{Joyeux:2011:rock}).
Precise control over the dynamic execution and interaction behavior of functional components, i.e. the computation and communication semantics, on model level
without hidden (i.e. code-defined) parts is urgently needed to enable the aforementioned separation of concerns between functional component development and
application-specific system integration.

Other existing model-based approaches beyond robotics such as OMG MARTE~\cite{MARTE}, AMALTHEA~\cite{Amalthea}, AADL~\cite{AADL:Introduction} and
SysML~\cite{SysML} offer concepts for describing the execution and interaction behavior of components on system level. However, central concepts are often
hidden in a \emph{freedom-of-choice} philosophy offering all kinds of alternative coequal concepts. Moreover, many of the concepts that are lifted to model
level are too fine granular (e.g. read and write operations on buffers in MARTE) directing the focus and efforts on minor aspects. In the end, the
\emph{robotics expert} who is mainly interested in functional development is left alone with many system-level design choices, while \emph{application domain
experts} need to understand all the low-level technical details (often on code level) of the functional components. This either leads to refusal of using the
model-based approaches in the first place, or results in non-interoperable, hard to reuse, functional components.

In this paper, we therefore pursue the opposed \emph{freedom-from-choice}~\cite{Lee:2010} approach by consciously restricting the modeling choices to the
crucial concepts and abstractions that are necessary to systematically design and integrate functional components.
More precisely, we present a meta-model using \emph{Model-Driven Software Engineering (MDSE)}, which is separated into two
parts, each individually addressing the corresponding concerns of the \emph{robotics experts} and the \emph{application domain experts}. The two
meta-model parts further allow to provide role-specific views with an appropriate abstraction level, and, they are interconnected, thus allowing to
ensure system level conformance by means of automated model consistency checks.

We pay special attention to non-functional system-level aspects such as an adequate responsiveness of
the overall system and deterministic system behavior.
As a core contribution in this paper we provide model-based mechanisms for \emph{robotics experts} to clearly define activation semantics for
the concurrent execution of functional blocks within components such that causal dependencies as well as data-flow characteristics (analogous to
SDF~\cite{Lee:1987:sdf}) between components are made explicit and can be consciously designed by \emph{application domain experts} on the right abstraction
level. Thereby, the abstraction level is chosen high enough to achieve separation of concerns between the corresponding developer roles, but also 
detailed enough to being able to calculate system-level end-to-end latencies and jitters for chains of interconnected components.

This paper is structured as follows. In the subsequent section \ref{sec:example}, we present a couple of real-world examples paying special attention to
non-functional, application related needs. Then, section \ref{sec:meta-model} presents a formal meta-model including a detailed explanation of the individual
core elements. Section \ref{sec:dsls} addresses possible model-editor syntax options based on the presented meta-model. Section \ref{sec:m2t} gives some
insights into the model-to-text transformations for two selected frameworks, namely ROS and \smart{}. Finally, section \ref{sec:related-works}
discusses related work, and section \ref{sec:conclusion} concludes the paper.

\section{Motivating Example}
\label{sec:example}

\begin{figure}[htb]
\begin{center}
  \includegraphics[width=1.0\columnwidth]{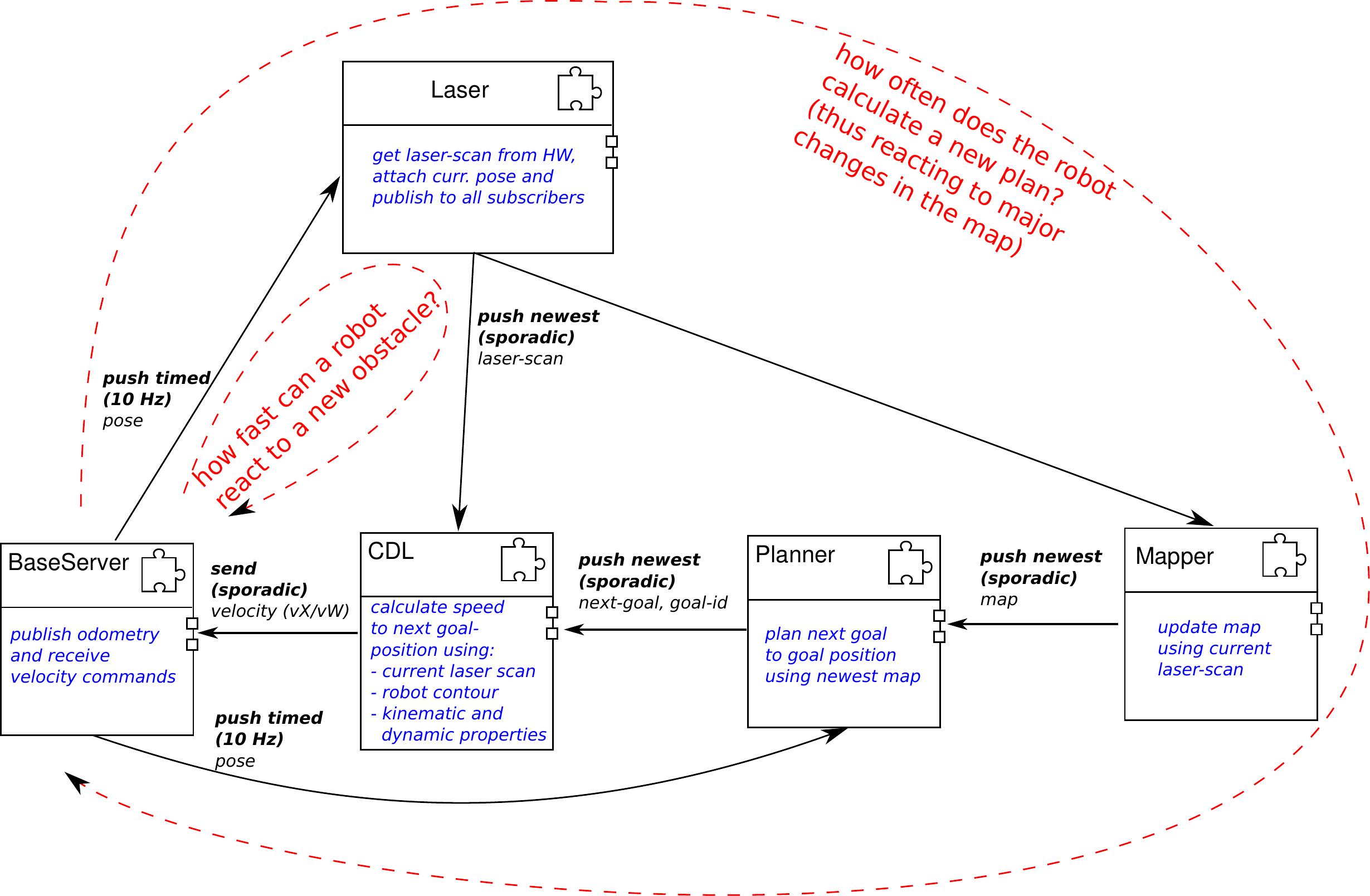}
  \caption{Data-flow (solid lines with arrow) between software components of the navigation scenario and two
  cause-effect chains addressing application-specific concerns (dashed lines with arrow)}
  \label{fig:navigation-scenario}
\end{center}
\end{figure}

This section presents a system example (see figure \ref{fig:navigation-scenario}) consisting of software components which have been used in various
real-world scenarios. This example represents a particular set of recurring robotics use-cases with an emphasis on application
specific, non-functional, system-level aspects.

Figure \ref{fig:navigation-scenario} presents the navigation scenario with two basic robot capabilities: the fast, local obstacle
avoidance (inner loop) and the slower grid-map-based path planning (outer loop). Each of these capabilities is realized by several connected components 
forming a component-chain between the involved sensors and actuators. The functional concerns are the described functionality, and,
in particular, the data exchange between the components.

A key non-functional requirement in this situation is the end-to-end response time for the 
obstacle avoidance. Concretely, if a human suddenly jumps in front of an autonomously navigating robot, 
how long will it take for the robot to react to this event by retrieving a new laser scan, 
propagating it to the obstacle avoidance component, calculating an evasive maneuver and finally commanding the robot-base? 
The maximum admissible value for this response time will probably influence
the periods at which individual components have to run, and possibly the choice of algorithms. 

Now, the selection of that value primarily depends on the kinematics constraints of the actual robot,
which may further be constrained by the concrete application (e.g. the acceptable maximum velocity of this robot moving in crowded areas). 
As these aspects are highly application specific, they need to remain
unbound until the corresponding \emph{application domain experts} provide the according domain knowledge allowing to select adequate values.

Another such sensor to actuator coupling (from here on we call it a \emph{cause-effect chain}) is the map-based path planning functionality. How often
does the path to a (remote) location need to be re-planned in order to adequately react to structural changes in the environment (such as closed or opened doors)? 
Again, depending on the expected environments the robot is supposed to operate in, the probability and frequency of changes can only be anticipated by 
\emph{application domain experts} providing the according requirements for the re-planning frequency.

Unfortunately, in the current state-of-practice, many details about the execution and interaction behavior are hidden within component implementations which
makes it difficult to reason about global execution aspects. 

For instance, in ROS the semantic of how incoming messages on a topic are processed is
intrinsically tied to the node implementation and cannot easily be changed according to a specific use-case.  
More precisely, the subscriber callback can directly process an incoming message or store it in a local variable (or buffer) 
for later processing in a timer callback (which is common pattern in ROS based systems). In the first case the
processing is data triggered, and the latter case corresponds to polling a register. Obviously, there are huge differences in the execution behavior with
respect to latency and jitter between both cases. We argue that the adequate execution and interaction behavior of functional components highly depends on the
actual application, and thus needs to remain a configurable part for \emph{application domain experts}. Furthermore, we are convinced that for systems comprising
many components this information needs to be lifted to model level to easily understand the overall system behavior.

There are many other comparable examples, e.g. related to tracking, person-following, human-robot interaction, visual servoing, etc. Interestingly enough, a
typical service robot, that combines multiple basic capabilities to achieve a certain task often needs to execute several such
\emph{cause-effect chains} in parallel, possibly with completely different requirements. These requirements can range from very strict hard real-time
guarantees (e.g. for a robot balancing on two wheels) up to very soft, safety unrelated, average timing estimations (e.g. reaction time in speech interaction).
Independent of the guarantee-severity (hard or soft), the important point addressed in this paper is that the mentioned system properties need to be
an explicit and adjustable part of the overall system design and \textbf{not} the result of hidden, too early, and unmodifiable decisions inside of component
implementations.

\section{Ecore Meta-Model for Component Definition and System Configuration}
\label{sec:meta-model}

\begin{figure*}[tb]
\begin{center}
  \includegraphics[width=1.0\textwidth]{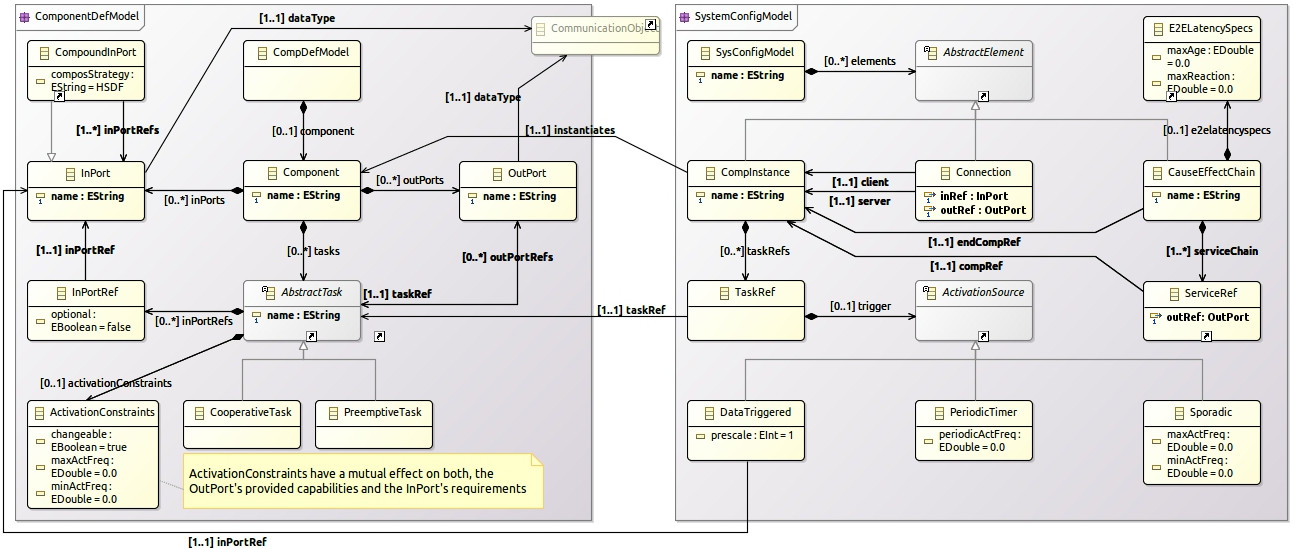}
  \caption{An Ecore meta-model, separating concerns in two model packages for component-definition (left) and system-configuration (right)}
  \label{fig:meta-model}
\end{center}
\end{figure*}

This section presents an ecore meta-model (shown in figure \ref{fig:meta-model}) which separates the individual concerns for component development and system
integration. It is worth noting that the presented meta-model is inspired by fully fledged robotics (meta)-models such as
\textsc{SmartMARS}~\cite{Schlegel:2012:inTech}. However, on the one hand, it has been simplified to focus on essential concepts for efficiency and clarity
reasons, and, on the other hand, it has been extended with additional concepts to address concerns related to inteconnected components with the
aforementioned \emph{cause-effect chains}.

There are several other component meta-models such as RobotML~\cite{Dhouib:2012:RobotML}, 
BRICS Component Model (BCM)~\cite{Bruyninckx:2013:BCM}, RTC~\cite{Ando:2005:RTC}, GCM from OMG 
MARTE~\cite{MARTE}, etc. which provide similar core items such as a \emph{Component}, an \emph{In-} and
\emph{OutPort}, an IDL for the definition of communicated data, a \emph{Connection} and often a \emph{ComponentInstance}. These meta-model
root-elements are not supposed to be reinvented here, instead, we encourage to map them onto the according original items of an already existing meta-model
wherever possible.

\subsection{Component-Definition Meta-Model}

The left part of figure \ref{fig:meta-model} addresses the \emph{robotics expert} view. The core element is the definition of a \emph{Component} including a
name, which serves as a unique identifier in the later component pool. The main purpose of a \emph{Component} is to provide clearly specified means of
communication (using \emph{In-} and \emph{OutPorts}) between the internal functionality (realized as \emph{Tasks}) and other components in the
system\footnote{Please note that the description of further important component's orchestration mechanisms such as component parametrization and the component's
lifecycle automaton~\cite{Lotz:2011:Monitoring} are considered out of scope in this paper.}.
\emph{In-}/\emph{OutPorts} represent typed, 1 to n, publish-subscribe communication semantics. In SOA terminology, an \emph{OutPort} is a publisher and resp. a
service provider, whereas an \emph{InPort} is a subscriber and resp. a service requestor. This communication semantics can be mapped onto many popular
middlewares such as the \emph{Data Distribution Service (DDS)} \cite{DDS}, onto other component models such as the Flow-Port from the OMG MARTE \cite{MARTE}
specification, or even directly onto existing (robotics) frameworks such as ROS, \smart{} and others. \emph{InPorts} additionally support the definition of
\emph{CompoundInPorts}, which allow to define advanced SDF~\cite{Lee:1987:sdf} composition strategies (such as HSDF). It is worth noting, that we do not (yet)
include other communication semantics such as request-response. There are many valid use-cases where components are only acting on request. However, such
components typically are not part of tightly coupled cause-effect chains between sensors and actuators, and thus can easily coexist alongside with the
extensions presented here.

At this point it is worth noting that we do not (yet) support hierarchical components (i.e. components of components). There are several approaches
in the robotics domain such as RTC~\cite{Ando:2005:RTC} or outside robotics such as MARTE~\cite{MARTE} providing hierarchical component models. In this paper we
chose a flat component representation focusing on the imminent problems first before generalizing and extending the model semantics. However, we might extend
our meta-model in future work accordingly.

\emph{Tasks} represent concurrent functionalities inside a \emph{Component} thereby clustering (independent) functional aspects, thus allowing to
implement more complex \emph{Components} that can provide several (independent) \emph{OutPorts}. This is particularly useful for sophisticated libraries such
as, for instance, OpenRAVE\footnote{OpenRAVE: \url{http://openrave.org/}}.
The main concern of a \emph{Task} is to (continuously) generate data for one or several \emph{OutPorts}.
Thereby, a \emph{Task} might internally use any kind of HW API (i.e. for sensors or actuators). For its computation a \emph{Task} can depend (strictly or
optionally) on data arriving from one or several \emph{InPorts}. Please note that \emph{InPorts} can be shared by several \emph{Tasks}. However, each
\emph{OutPort} must be served by exactly one distinct \emph{Task}. This is an important aspect for the configuration of interconnected components in
\emph{cause-effect chains} (see below).
Furthermore, two different \emph{Task}-types are distinguished. \emph{PreemptiveTasks} can be executed in parallel (allowing e.g. to utilize multicore CPUs).
In case that \emph{PreemptiveTasks} share data inside a component, it is assumed that the component developer implements suitable mutual exclusion mechanisms.
\emph{CooperativeTasks} are executed pseudo-parallel (they are internally sequenced), thus preventing race-conditions
even if accessing unprotected shared data.

One of the particularly interesting aspects is the optional definition of \emph{ActivationConstraints}. \emph{ActivationConstraints} are used to express
intrinsic requirements on activation characteristics of a \emph{Task}. Application-specific activation characteristics (such as configurable timers, or the
synchronicity of data received on one or several \emph{InPorts}) should be left open for later configuration by the \emph{application domain expert} who is
responsible for system integration (see subsection~\ref{sec:sys_config}).
\emph{ActivationConstraints} can be used by \emph{robotics experts} though, for instance, to express strict and unmodifiable constraints on execution
characteristics, which might be due, for instance, to an internal HW trigger (e.g. a sensor providing data with a fixed frequency). Other use-cases for
specifying \emph{ActivationConstraints} during component development include internally used algorithms  requiring specific activation-frequency ranges (e.g.
for a PID controller).

\subsection{System Configuration Meta-Model}
\label{sec:sys_config}

The right part of figure \ref{fig:meta-model} addresses the \emph{application domain expert's} view. The main concern here is the initialization of
\emph{ComponentInstances} and the definition of initial \emph{Connections}\footnote{Some frameworks such as \smart{} additionally allow for dynamic (re-)wiring
at run-time, which is out of scope in this discussion.} between \emph{In-} and \emph{OutPorts}. In future work we plan to link the system configuration model
with a deployment model (such as e.g. in \cite{Schlegel:2012:inTech}) and a simple platform definition model which altogether embody the overall
system integration step.

The novel parts in the meta-model are the late binding of the \emph{ActivationSource} for each corresponding \emph{Task} reference and the definition of
\emph{CauseEffectChains}.
There is an interesting interdependency between the specification of \emph{ActivationConstraints} in the component definition model and the corresponding
selection of an \emph{ActivationSource} in the system configuration model. The \emph{ActivationSource} enables \emph{application domain experts} to select
specific execution characteristics for a \emph{Task} considering the predefined boundaries in the according \emph{ActivationConstraints}. 

There are three different types of \emph{ActivationSources}. One is the \emph{DataTrigger} denoting that each incoming data message on the referenced
\emph{InPort} directly triggers the execution of the associated \emph{Task}. In other words, the \emph{Task} synchronously reads data from the referenced
\emph{InPorts}. By definition, we allow at most for one \emph{InPort} with \emph{DataTrigger} semantics per \emph{Task}. All other \emph{InPorts} are
asynchronously read with register semantics at the time instant of the \emph{Task's} activation. Please note, that by using \emph{CompoundInPorts} it is
possible to define more complex data triggered activation schemes involving several \emph{InPorts} (such as homogeneous SDF~\cite{Lee:1987:sdf}, or respectively
AND- and OR-activation semantics \cite{symtas}).

Another option is to use a \emph{PeriodicTimer} as \emph{ActivationSource} for a \emph{Task}. In this case all referenced \emph{InPorts} are asynchronously read
with register semantics at periodic \emph{Task} activation.

The third \emph{ActivationSource} type is \emph{Sporadic}. Its main use-case is the backwards compatibility for already implemented components whose hard-coded
\emph{Task} activation behavior is unmodifiable on model level.

Next, a \emph{CauseEffectChain} is defined by a list of \emph{OutPort} references. This is sufficient to unambiguously derive all relevant model elements
contained in the cause-effect chain, namely the \emph{Connections}, the \emph{InPort} to \emph{Task} dependencies as well as the \emph{Task} to \emph{OutPort} dependencies.
One of the main concerns of the \emph{CauseEffectChain} is to define a relationship between the overall \emph{E2E\-La\-ten\-cy\-Specs} and the involved
\emph{ActivationSources}. The individual \emph{ActivationSources} can be chosen such that both requirements are satisfied, those coming from the enclosing
component (defined by the according \emph{ActivationConstraints}) and those coming from the involved \emph{CauseEffectChain} (defined by the corresponding
\emph{E2E\-La\-ten\-cy\-Specs}).

The main advantage now is, that application domain specific requirements (such as the reaction time for evading an obstacle) can be directly annotated by the
\emph{E2E\-La\-ten\-cy\-Specs} individually for each \emph{CauseEffectChain}. Furthermore, these requirements are directly mapped to according configurations
within individual components. The exact run-time characteristics depend on the actually used platform. For instance, a platform providing a real-time scheduler
and real-time communication can be configured to exactly meet the execution requirements. Another possibility is a regular best-effort system which can be
configured to meet average run-time characteristics, accepting rare time violations which can be detected and handled accordingly at run-time using generated
watchdogs and other monitoring techniques (such as \cite{Lotz:2011:Monitoring}).

\section{DSLs and Model Checks}
\label{sec:dsls}

\begin{figure}[htb]
\begin{center}
  \includegraphics[width=1.0\columnwidth]{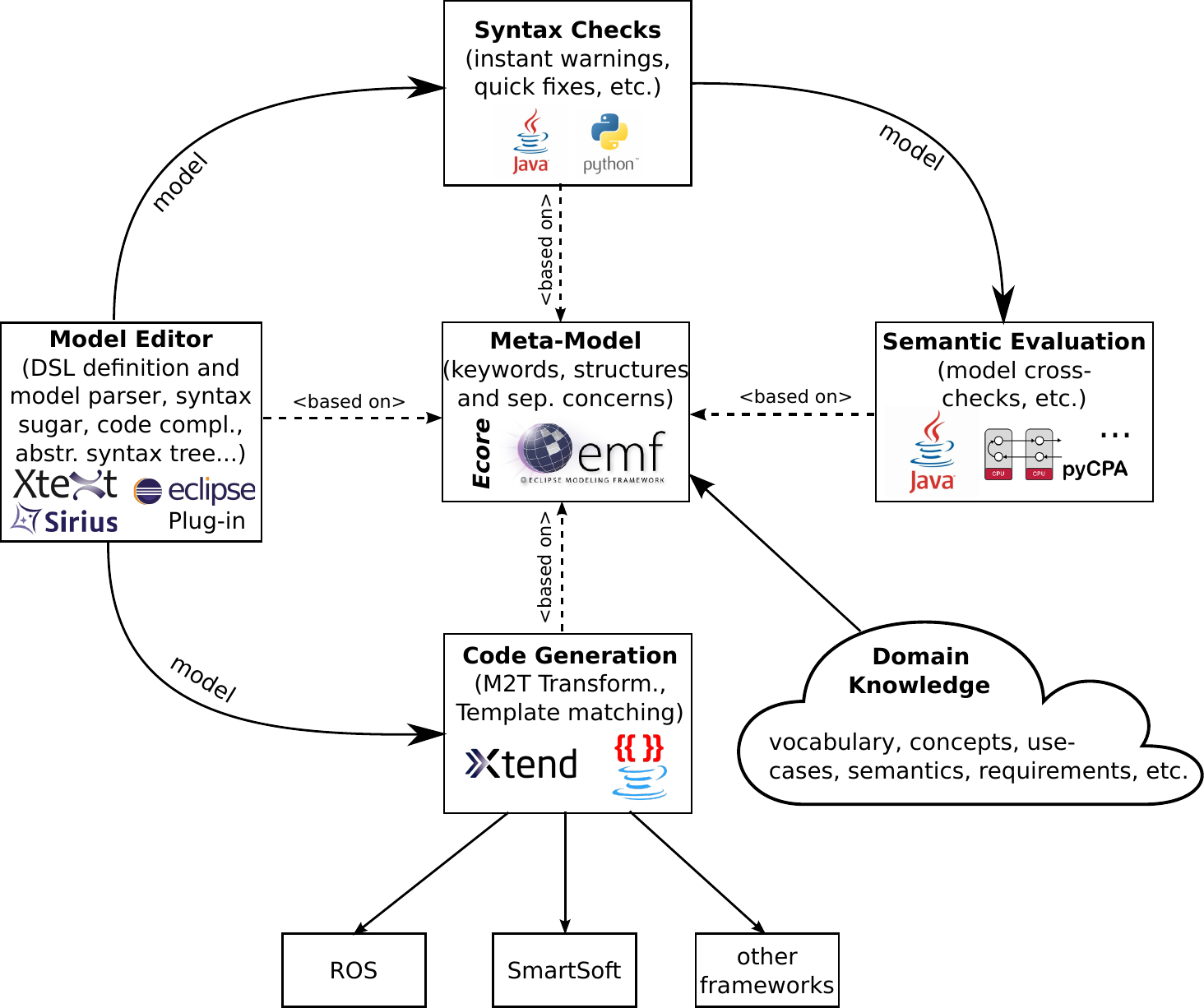}
  \caption{Common modeling tool-artifacts involved in the realization of a modeling approach}
  \label{fig:modelling-overview}
\end{center}
\end{figure}

A meta-model as presented in the previous section is the foundation in a modelling initiative which formalises domain knowledge by specifying domain specific
vocabulary with the involved structures and relations. In addition, further aspects such as an intuitive model editor with easily understandable
representation, as well as an unambiguous mapping into code by model-to-text transformations (see overview in figure \ref{fig:modelling-overview})
are important factors for the overall acceptance and usefulness of a modeling tool.

This section addresses issues related to the design and development of a model editor based on the meta-model presented in section \ref{sec:meta-model}.
One of the first steps is the decision for a graphical or a textual representation. Both have their strengths and weaknesses (as shown in the following
two subsections). It is sometimes even possible to combine or mix both representations by e.g. providing a textual editor and generating
a graphical representation on the fly, or by providing a graphical editor including elements with text fields embedding a textual model editor. In the
end, the final decision often is a matter of individual preferences in the according domain for which the tool is designed.

The following two subsections individually present excerpts of the model editors according to the two EPackages defined in the meta-model in figure
\ref{fig:meta-model}. One is a graphical DSL for component definition and the other is a textual DSL for system configuration.

\subsection{Component-Definition Model}

\begin{figure}[htb]
\begin{center}
  \includegraphics[width=0.35\columnwidth]{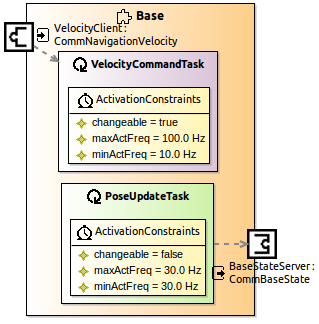}
  \includegraphics[width=0.62\columnwidth]{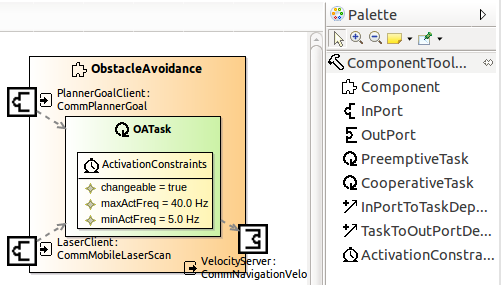}
  \caption{Exemplary graphical DSL notation demonstrated using the components: \emph{Base} (left) and \emph{ObstacleAvoidance} (right)}
  \label{fig:graphical-model}
\end{center}
\end{figure}

This subsection presents an exemplary graphical notation for the component definition according to the EPackage \emph{ComponentDefModel} in figure
\ref{fig:meta-model} which is the view of \emph{robotics experts}. The model editor is demonstrated using the \emph{ObstacleAvoidance} and the \emph{Base}
components (see figure \ref{fig:graphical-model}). The graphical editor is based on the Eclipse Sirius\footnote{\url{www.eclipse.org/sirius/}} plugin whose 
graphical notation is inspired by UML. In fact, another possibility is to directly profile UML as e.g. demonstrated by
RobotML~\cite{Dhouib:2012:RobotML}.

The interesting parts are the dashed lines between \emph{InPorts} and the \emph{Tasks} as well as between the \emph{Tasks} and the \emph{OutPorts}. This way,
the functional dependency of a \emph{Task} to input data as well as the responsibility of that \emph{Task} to provide results on a certain \emph{OutPort} are
clearly specified. This allows to implement functionally complete and compilable code without already binding the exact run-time communication semantics.

In case the \emph{changeable} flag is set to false (see e.g. \emph{PoseUpdateTask} in figure \ref{fig:graphical-model}), this indicates that the provided
\emph{ActivationConstraints} are fix and can not be changed any more during system configuration. This way it is possible to express strict requirements which
should not be changed by \emph{application domain experts}. If in addition, both values of \emph{maxActFeq} and \emph{minActFreq} are equal, this indicates a
fix (i.e. unmodifiable), \emph{periodic} update frequency.

\subsection{System Configuration Model}

\begin{figure}[htb]
\begin{center}
  \includegraphics[width=1.0\columnwidth]{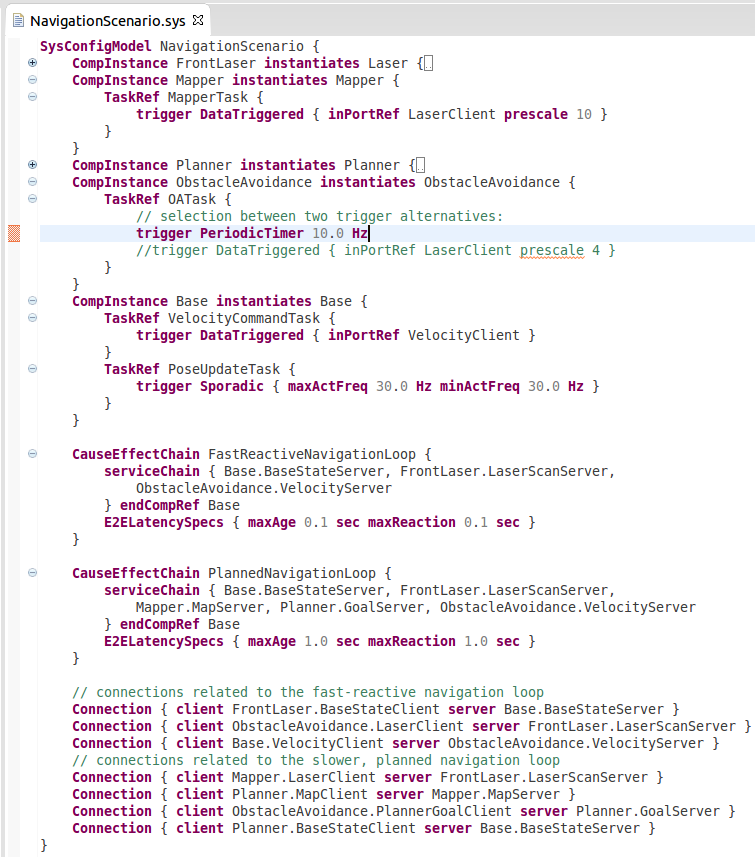}
  \caption{Navigation-scenario model using an Xtext based DSL}
  \label{fig:nav-model}
\end{center}
\end{figure}

Figure \ref{fig:nav-model} shows an excerpt of the navigation scenario (presented in section \ref{sec:example}) using an Xtext based DSL according
to the EPackage \emph{SystemConfigModel} in figure \ref{fig:meta-model} which defines the high level view for \emph{application domain experts}.

For the presented DSL, there are several factors in favor for a textual representation rather than graphical. For instance, the model in figure
\ref{fig:nav-model} references lots of already existing elements: \emph{Components}, \emph{Tasks} and \emph{In-/OutPorts}. Therefore, Xtext allows to implement
powerful code completion mechanisms using scope providers and content assists to e.g. generate higher level model elements including their child elements (such
as a \emph{ComponentInstance} with its \emph{TaskRefs}).

Furthermore, the definition of a \emph{CauseEffectChain} is mainly based on a list of concatenated
\emph{OutPort} references. A scope provider in combination with a validation check ensures that only those successive \emph{OutPort}
references can be chosen which really are reachable from the current \emph{OutPort} reference in the list (through according \emph{Connection} and the
dependency specifications within the corresponding component). Furthermore, as there might be lots of involved components (20 and more) in a typical system,
graphical notations tend to become cumbersome with lots of crossing lines for e.g. the \emph{Connections}. Even so, textual representations can also get
lengthy, it still is easier to distribute a textual model over several files.

\begin{figure}[htp]
\begin{center}
  \includegraphics[width=1.0\columnwidth]{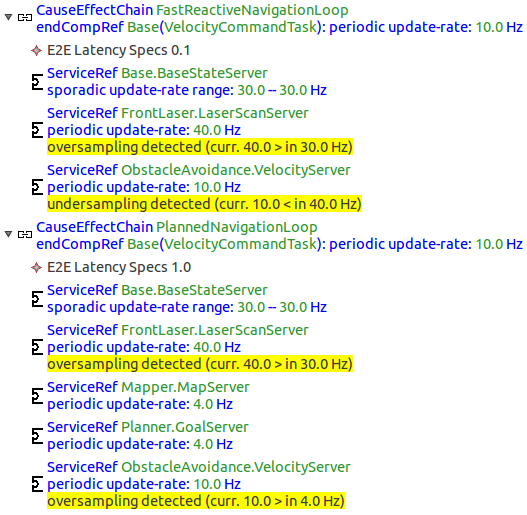}
  \caption{Navigation-scenario on-line model-evaluation results including activation frequency propagation and over-/under-sampling checks}
  \label{fig:model-eval}
\end{center}
\end{figure}

A model-editor additionally supports the editing process by on-the-fly checking the syntax according to the meta-model specification and by additionally running
semantic evaluation checks (see figure \ref{fig:modelling-overview}). For the former, Xtext allows to implement element-based \emph{Validation Rules} which
display error- or warning-messages attached to a corresponding textual element in the editor. For the latter semantic evaluation, there are two main realization
possibilities. One is to use an external 3rd-party tool such as e.g. \emph{pyCPA}~\cite{pyCPA} or \emph{SymTA/S}~\cite{symtas}, for analyzing worst-case
latencies along cause-effect chains whose input can be directly generated from the Xtext model (we plan to demonstrate this in our future work).
Another option is to directly implement semantic interpretation rules as part of the model editor and to
display their results in the \emph{Outline} view of the Xtext model (as shown in figure \ref{fig:model-eval}).

For example, the \emph{FastReactiveNavigationLoop} in figure \ref{fig:nav-model} defines a cause-effect chain consisting of four components: (1) a \emph{Base}
component providing odometry, (2) a \emph{Laser} component providing laser-scans, (3) an \emph{ObstacleAvoidance} component and (4) again the \emph{Base}
component receiving velocity commands. For each \emph{Task} within these components an individual \emph{ActivationSource} can be selected.
This way, the \emph{Tasks} form a concatenated chain whose links are either synchronously connected using the \emph{DataTriggered} activation source or
asynchronously connected using e.g. the \emph{TimedTrigger} activation source. Thereby, all successive \emph{Tasks} with the \emph{DataTriggered} activation
source implicitly follow the update frequency from the preceding \emph{Task}. Along the chain, this frequency can be subdivided by an optional \emph{prescaler}
as is demonstrated by the \emph{Mapper} component in figure \ref{fig:model-eval} subdividing the incoming frequency of 40 Hz from the
\emph{Laser} component by 10 in order to get an adequate frequency of 4 Hz. It is worth noting that since there is a 1-to-1 relationship between \emph{Tasks}
and \emph{OutPorts} one might be tempted to combine both modeling elements, e.g. by including the \emph{Task} semantics into the \emph{OutPort}. However, a
\emph{Task} additionally serves as a functional block for the \emph{InPorts} which might be independent from the \emph{OutPort} as e.g. is demonstrated by the
\emph{Base} component in figure \ref{fig:graphical-model}.

\emph{Tasks} in a chain using \emph{PeriodicTimers} can either run at a higher update frequency than the input data, thus potentially using old values for
several task-cycles (see \emph{oversampling} in figure \ref{fig:model-eval}), or at a lower update frequency, thus skipping some intermediate values (see
\emph{undersampling} in figure \ref{fig:model-eval}). Depending on the current application, \emph{over-} or \emph{under-sampling} might be acceptable or not.
The important point is that this information is available for the \emph{application domain expert}, thus enabling him to find the right balance between the
different selection options of the individual \emph{ActivationSources}. For example, the expert could decide to use a \emph{DataTriggered} activation source
with prescale 4 instead of the \emph{PeriodicTimer} for the \emph{OATask} in the \emph{ObstacleAvoidance} component (see figure \ref{fig:nav-model}) . This
would result in a triggering of the \emph{OATask} each 4th incoming laser-scan, thus again getting an update-frequency of 10 Hz, however, now synchronously
without sampling effects due to scheduling.

\section{M2T Code-Generation}
\label{sec:m2t}

One of the remaining elements in figure \ref{fig:modelling-overview} which is not yet discussed is the model-to-text transformation (i.e. the code
generation). Model-to-text transformations implement the actual grounding of the meta-model into the code. At the moment, lots of valuable algorithms for
robotics are implemented as libraries (often embedded in ROS nodes or enveloped by \smart{} components). Therefore, we consider it illusory (at least in the
near future) to describe all necessary low level details in an overall model, and to completely generate ready to use components by simply pushing a button.
Instead, we focus on modeling essential parts related to structured system integration and generate glue-code (e.g. using the generation-gap pattern)
to link with existing implementations.

\begin{figure}[htb]
\begin{center}
  \includegraphics[width=0.8\columnwidth]{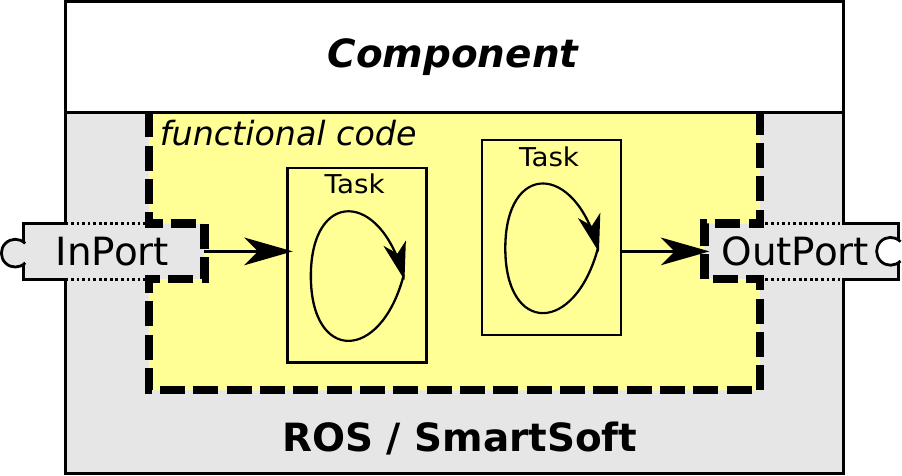}
  \caption{The code generation depends on a generic interface (dashed line) between the functional code (component's inner area) and the framework glue-code
  (component's gray container).}
  \label{fig:code-gen}
\end{center}
\end{figure}

Successful code generation heavily relies on a generic interface (see figure \ref{fig:code-gen}) between the generated framework glue-code (e.g. for ROS or
\smart{}) and the provided functional code. It is a matter of framework capabilities whether the glue code is generated from the beginning during component
design providing according configuration options (e.g. using the parameter specification in \smart{}), or whether the glue-code is afterwards generated based on
the system configuration model as it is typically the case for ROS nodes. In any case, the main concern for code generation is to preserve model semantics with
respect to the designed execution and communication behavior.
 
In order to ease the migration of already existing components, we support both, \emph{top-down} and \emph{bottom-up} development. 
\emph{Top-down} refers to designing new components which can be adjusted during system configuration on model level without modifying their functional
code. \emph{Bottom-up} refers to existing component implementations where we express their implemented execution and communication behavior with our model.

\section{Related Work}
\label{sec:related-works}

In the last decade, component-based frameworks for robotics have become the norm. 
They mostly focus on implementing functional blocks and abstracting over
communication middlewares.
However, as argued in the introduction, structured system integration allowing to precisely control 
the dynamic execution and interaction behavior of functional components on model level, according to application specific needs and 
without hidden code-defined parts is one of the hot research topics.

Some initial works within the robotics domain address parts of the aforementioned problem. For instance, systematic
component development and structured system integration relies on a clear separation of concerns as is also recognized in the BRICS project as the
5Cs~\cite{Bruyninckx:2013:BCM} (computation, communication, configuration, coordination and composition). Separating concerns means to systematically
structure the model representations, as e.g. demonstrated in RobotML~\cite{Dhouib:2012:RobotML}, by separating models in packages related to communication,
behavior, architecture, and deployment. Precise concepts for addressing these concerns are, however, still under discussion and we see our 
activation semantics as a concrete contribution in this direction.

An existing proposal beyond robotics to describe such properties is provided by OMG MARTE~\cite{MARTE} 
on a very detailed model level. 
However, we argue that it is exactly this detailed level that makes it too complex for practical usage.
Specifically, MARTE's \emph{Generic Component Model} (GCM) offers \emph{flow-ports} and 
\emph{client-server ports}. Flow-ports are taken from SysML, and are distinguished into
``push'' and ``pull'' ports (cf.~\cite[section 12.3]{MARTE}), which can be mapped to our
DataTriggered and TimeTriggered (assuming that the ``pull'' is caused by a timer) 
semantics, respectively. A first issue is that for ``pull'' ports,
the desired timing is not clearly marked as such, but can only be derived from a model
which directly describes implementation concepts such as OS-specific alarms (cf.~\cite[section 
14.1.6.2]{MARTE}). Similarly,  client-server ports are described using behavior state machines, 
whose analysis poses a hard problem. In contrast, we propose a clearly marked ``PeriodicTimer'' 
activation concept that is independent from a concrete implementation and facilitates much more 
straightforward analysis. Thus, the proposed model is on a higher abstraction level than MARTE.

We are also not aware of any concrete implementation for translation of models using MARTE 
to real-world robotics frameworks, which may certainly be due to its complexity. However, in order
for a modeling approach to be accepted and used, models need to be supported by 
MDE tools, and be integrated with commonly used robotic frameworks. Our approach has, so far, been 
applied to both ROS~\cite{Quigley:2009:ROS} and \smart{}~\cite{Schlegel:2012:inTech}.

An approach closer to our concepts can be found in the \emph{Architecture Analysis and Design
Language} (AADL) \cite{AADL:Introduction}. In particular, \cite{AADL:Flows} describes how \emph{flows} in AADL can be
used to model activation semantics similar to ours, using appropriate port types (i.e. queued or sampled),
and thread types (aperiodic with trigger, for \emph{DataTriggered} semantics, and periodic for \emph{TimeTriggered}).
It also supports attributes such as \emph{deadlines} for end-to-end latency analysis using the OSATE2
tool, as also demonstrated for robotics in \cite{Biggs:2014:AADL}. 

Furthermore, AADL allows modeling many additional orthogonal aspects, e.g. the functional behavior inside components using threads, function calls, etc., or
details of the execution platform consisting of data buses, CPUs, etc. While this renders AADL very powerful, it is left open how these different concepts
addressing different concerns are to be used for system design. As a consequence, the average user might have difficulties to adequately use AADL which
threatens its practical usefulness. According to our experience, smaller models that focus on a coherent set of system engineering concerns are of higher
practical usefulness since they are far more comprehensive. Ultimately, coherency along with simplicity is key to practically achieve separation of concerns in
model-based design, and thus  to cope with system complexity.

The only \emph{robotics} initiative that we are aware of following a similar ``freedom from choice''
approach is the ``oroGen'' tool from the Robot Construction Kit (Rock)~\cite{Joyeux:2011:rock}.
In particular, it distinguishes time and data-triggered activation of
components\footnote{\url{rock-robotics.org/stable/documentation/orogen/triggering/index.html}}.
This is a pre-requisite for a precise analysis, but effect chains as such are not provided by oroGen. 
The other concepts are compatible, however, so they would certainly be a straightforward addition to 
oroGen's underlying component model.

Finally, we would like to note that formally specified activation semantics can also be supported 
in a ``freedom of choice'' model, of course. Examples include UML's port behavior state machines,
or RTC's execution semantics \cite{Ando:2005:RTC}. However, our approach differs in that we intentionally
limit the modeling choices to a small set of activation semantics sufficient to ease integration 
and checking.

\section{Conclusions and Future Works}
\label{sec:conclusion}

In recent years robotics technologies have become an integral part of everyday life, sometimes embedded in products such as car driving-assistants
and sometimes more apparent as smart home-cleaning devices. The public expectations for future robotics technologies are high. 
Robotics research fosters these expectations by presenting impressive lab prototypes, yet, until now only a few (rather simple) examples have been realized as
products. We believe that one of the main reasons is a general lack of appropriate software engineering methods for systematic integration
allowing to cope with the vast complexity as is common in autonomous robotic systems.

This paper provides a meta-model which clearly separates different concerns from
component developers and system integrators enabling them to collaboratively design and develop component-based robotic software systems. This separation of
concerns is achieved: (i) by enabling component developers to focus their engineering efforts on functional concerns, without
presuming any system-level application-specific details, and (ii) by enabling system integrators to fully understand and adjust the execution and communication semantics
of components on model level according to application-specific requirements, without the need to investigate or adapt internal implementations.

We thereby carefully balanced between the \emph{freedom-of-choice} and \emph{freedom-from-choice} \cite{Lee:2010} philosophies by providing as much
design-freedom as possible for the individual developers while restricting their design-choices where needed to ensure interoperability, reusability, and 
overall system consistency. 

To this effect, the presented work provides a contribution for making the step from function-driven system-level coding towards structured, application-specific, and
model-driven system integration.

\addtolength{\textheight}{-17cm}   


\bibliographystyle{IEEEtran}
\bibliography{references}

\begin{thebibliography}{10}
\providecommand{\url}[1]{#1}
\csname url@samestyle\endcsname
\providecommand{\newblock}{\relax}
\providecommand{\bibinfo}[2]{#2}
\providecommand{\BIBentrySTDinterwordspacing}{\spaceskip=0pt\relax}
\providecommand{\BIBentryALTinterwordstretchfactor}{4}
\providecommand{\BIBentryALTinterwordspacing}{\spaceskip=\fontdimen2\font plus
\BIBentryALTinterwordstretchfactor\fontdimen3\font minus
  \fontdimen4\font\relax}
\providecommand{\BIBforeignlanguage}[2]{{%
\expandafter\ifx\csname l@#1\endcsname\relax
\typeout{** WARNING: IEEEtran.bst: No hyphenation pattern has been}%
\typeout{** loaded for the language `#1'. Using the pattern for}%
\typeout{** the default language instead.}%
\else
\language=\csname l@#1\endcsname
\fi
#2}}
\providecommand{\BIBdecl}{\relax}
\BIBdecl

\bibitem{SPARC}
SPARC, ``{European SPARC Robotics Initiative},'' Online,
  http://sparc-robotics.eu/.

\bibitem{MAR}
\BIBentryALTinterwordspacing
``{Robotics 2020 Multi-Annual Roadmap},'' {SPARC Robotics -- The Partnership
  for Robotics in Europe}, February 2015, call 2 ICT24 -- Horizon 2020.
  [Online]. Available:
  \url{http://www.eu-robotics.net/cms/upload/Multi-Annual_Roadmap2020_ICT-24_R%
ev_B_full.pdf}
\BIBentrySTDinterwordspacing

\bibitem{Lotz:2014:IJISMD}
A.~Lotz, J.~F. Ingl\'{e}s-Romero, D.~Stampfer, M.~Lutz, C.~Vicente-Chicote, and
  C.~Schlegel, ``{Towards a Stepwise Variability Management Process for Complex
  Systems: A Robotics Perspective},'' in \emph{International Journal of
  Information System Modeling and Design (IJISMD)}, vol.~5, no.~3.\hskip 1em
  plus 0.5em minus 0.4em\relax IGI Global, 2014, pp. 55--74.

\bibitem{Ando:2005:RTC}
N.~Ando, T.~Suehiro, K.~Kitagaki, T.~Kotoku, and W.-K. Yoon, ``Rt-component
  object model in rt-middleware distributed component middleware for rt (robot
  technology),'' in \emph{IEEE International Symposium on Computational
  Intelligence in Robotics and Automation (CIRA '05)}, June 2005, pp. 457
  --462.

\bibitem{Dhouib:2012:RobotML}
S.~Dhouib, S.~Kchir, S.~Stinckwich, T.~Ziadi, and M.~Ziane, ``{RobotML}, a
  {D}omain-{S}pecific {L}anguage to {D}esign, {S}imulate and {D}eploy {R}obotic
  {A}pplications,'' in \emph{3rd {I}nt. {C}onf. on {S}imulation, {M}odeling and
  {P}rogramming for {A}utonomous {R}obots ({SIMPAR})}, ser. Lecture Notes in
  Computer Science, I.~Noda, N.~Ando, D.~Brugali, and J.~Kuffner, Eds.\hskip
  1em plus 0.5em minus 0.4em\relax Springer Berlin Heidelberg, November 2012,
  vol. 7628, pp. 149--160.

\bibitem{Bruyninckx:2013:BCM}
H.~Bruyninckx, M.~Klotzb\"{u}cher, N.~Hochgeschwender, G.~Kraetzschmar,
  L.~Gherardi, and D.~Brugali, ``{The BRICS component model: a model-based
  development paradigm for complex robotics software systems},'' in \emph{Proc.
  of the 28th Annual ACM Symposium on Applied Computing}, ser. SAC '13.\hskip
  1em plus 0.5em minus 0.4em\relax New York, NY, USA: ACM, 2013.

\bibitem{Joyeux:2011:rock}
S.~Joyeux and J.~Albiez, ``Robot development: from components to systems,'' in
  \emph{6th National Conference on Control Architectures of Robots}, Grenoble,
  France, May 2011.

\bibitem{MARTE}
\BIBentryALTinterwordspacing
MARTE, ``{A UML Profile for MARTE: Modeling and Analysis of Real-Time Embedded
  systems},'' June 2011, v 1.1. [Online]. Available:
  \url{http://www.omg.org/spec/MARTE/1.1/}
\BIBentrySTDinterwordspacing

\bibitem{Amalthea}
\BIBentryALTinterwordspacing
``Amalthea -- an open platform project for embedded multicore systems,''
  checked on August 7th, 2015. [Online]. Available:
  \url{http://www.amalthea-project.org}
\BIBentrySTDinterwordspacing

\bibitem{AADL:Introduction}
P.~H. Feiler, D.~P. Gluch, and J.~J. Hudak, ``{The Architecture Analysis and
  Design Language: An Introduction},'' Carnegie Mellon University, Technical
  Note CMU/SEI-2006-TN-011, 2006.

\bibitem{SysML}
\BIBentryALTinterwordspacing
``{OMG Systems Modeling Language (SysML)},'' {Object Management Group (OMG)},
  June 2012, v1.3. [Online]. Available:
  \url{http://www.omg.org/spec/SysML/1.3/}
\BIBentrySTDinterwordspacing

\bibitem{Lee:2010}
E.~A. Lee, ``{Disciplined Heterogeneous Modeling},'' in \emph{MODELS 2010},
  Oslo, Norway, October 2010, invited Keynote Talk.

\bibitem{Lee:1987:sdf}
E.~Lee and D.~Messerschmitt, ``Synchronous data flow,'' \emph{Proceedings of
  the IEEE}, vol.~75, no.~9, pp. 1235--1245, Sept 1987.

\bibitem{Schlegel:2012:inTech}
C.~Schlegel, A.~Steck, and A.~Lotz, ``Robotic software systems: From
  code-driven to model-driven software development,'' in \emph{Robotic Systems
  - Applications, Control and Programming}, A.~Dutta, Ed.\hskip 1em plus 0.5em
  minus 0.4em\relax InTech, 2012, pp. 473--502, {ISBN: 978-953-307-941-7}.

\bibitem{Lotz:2011:Monitoring}
A.~Lotz, A.~Steck, and C.~Schlegel, ``{Runtime Monitoring of Robotics Software
  Components: Increasing Robustness of Service Robotic Systems},'' in
  \emph{15th International Conference on Advanced Robotics (ICAR)}, Tallinn
  (Estonia), June 2011.

\bibitem{DDS}
DDS, ``{OMG Data Distribution Service for Real-time Systems},'' OMG Document
  formal/07-01-01, January 2007, version 1.2.

\bibitem{symtas}
R.~Henia, A.~Hamann, M.~Jersak, R.~Racu, K.~Richter, and R.~Ernst, ``{System
  level performance analysis - the SymTA/S approach},'' \emph{Computers and
  Digital Techniques, IEE Proceedings -}, vol. 152, no.~2, pp. 148--166, Mar
  2005.

\bibitem{pyCPA}
J.~Diemer, P.~Axer, and R.~Ernst, ``{Compositional Performance Analysis in
  Python with pyCPA },'' in \emph{3rd Int. Workshop on Analysis Tools and
  Methodologies for Embedded and Real-time Systems (WATERS)}, 2012.

\bibitem{Quigley:2009:ROS}
M.~Quigley, B.~Gerkey, K.~Conley, J.~Faust, T.~Foote, J.~Leibs, E.~Berger,
  R.~Wheeler, and A.~Ng, ``{ROS: an open-source Robot Operating System},''
  \emph{ICRA Workshop on Open Source Software}, 2009.

\bibitem{AADL:Flows}
P.~Feiler and J.~Hansson, ``{Flow Latency Analysis with the Architecture
  Analysis and Design Language (AADL)},'' Carnegie Mellon University, Technical
  Note CMU/SEI-2007-TN-010, 2007.

\bibitem{Biggs:2014:AADL}
G.~Biggs, K.~Fujiwara, and K.~Anada, ``Modelling and analysis of a redundant
  mobile robot architecture using aadl,'' in \emph{Simulation, Modeling, and
  Programming for Autonomous Robots (SIMPAR 2014)}.\hskip 1em plus 0.5em minus
  0.4em\relax Springer International Publishing, 2014, pp. 146--157.

\end{thebibliography}



\end{document}